\documentclass[conference]{IEEEtran}
\IEEEoverridecommandlockouts
\usepackage{cite}
\usepackage{amsmath,amssymb,amsfonts}
\usepackage{algorithmic}
\usepackage{graphicx}
\usepackage{textcomp}
\usepackage{xcolor}
\usepackage{url}

\usepackage{algorithm}
\usepackage{verbatim}
\usepackage{listings}
\usepackage{pifont}

\newcommand{\add}[1]{#1}
\newcommand{\addvone}[1]{#1}
\newcommand{\addvtwo}[1]{#1}

\def\BibTeX{{\rm B\kern-.05em{\sc i\kern-.025em b}\kern-.08em
    T\kern-.1667em\lower.7ex\hbox{E}\kern-.125emX}}

\begin{document}

\title{GNNHLS: Evaluating Graph Neural Network Inference via High-Level Synthesis\thanks{This manuscript has been subsequently published~\cite{zdczc23}.}}

\author{\IEEEauthorblockN{Chenfeng Zhao, Zehao Dong, Yixin Chen, Xuan Zhang, Roger D. Chamberlain}
\IEEEauthorblockA{\textit{McKelvey School of Engineering} \\
\textit{Washington Univ. in St. Louis}\\
\{chenfeng.zhao,zehao.dong,ychen25,xuan.zhang,roger\}@wustl.edu}
}

\maketitle

\begin{abstract}
With the ever-growing popularity of Graph Neural Networks (GNNs), 
efficient GNN inference is gaining tremendous attention. 
Field-Programming Gate Arrays (FPGAs) are a promising execution platform 
due to their fine-grained parallelism, low-power consumption, reconfigurability, and concurrent execution. Even better, High-Level Synthesis (HLS) tools bridge the gap between the non-trivial FPGA development efforts and rapid emergence of new GNN models.
In this paper, we propose GNNHLS, an open-source framework to comprehensively evaluate GNN inference acceleration on FPGAs via HLS, containing a software stack for data generation and baseline deployment, and FPGA implementations of 6 well-tuned GNN HLS kernels. 
We evaluate GNNHLS on 4 graph datasets with distinct topologies and scales. The results show that GNNHLS achieves up to $50.8\times$ speedup and $423\times$ energy reduction relative to the CPU baselines. Compared with the GPU baselines, GNNHLS achieves up to $5.16\times$ speedup and $74.5\times$ energy reduction.
\end{abstract}


\section{Introduction}

Graphs are widely adopted to model the relational-structured data in social networks, bioinformatics, etc~\cite{zhao2023supercut}. 
Machine learning (ML) on graphs has experienced a surge of popularity in the past decade, since
traditional ML models, which are designed to process Euclidean data with regular structures, are ineffective at performing prediction tasks on graphs.
Due to their simplicity and superior representation learning ability, Graph Neural Networks (GNNs)~\cite{kipf2016semi,velivckovic2017graph,xu2018how,zhang2018end,dong2023cktgnn} 
have achieved impressive performance on various graph learning tasks, such as node classification, graph classification, etc. 


To implement 
GNNs, a set of widespread libraries, such as PyTorch Geometric (PYG)~\cite{fey2019fast} and Deep Graph Library (DGL)~\cite{wang2019deep}, are built upon general-purpose ML frameworks (e.g. PyTorch~\cite{paszke2019pytorch}) targeting CPU and GPU platforms. 
However, the performance and energy consumption of GNN implementations are hindered by 
both hardware platforms and software frameworks:
(1)~Distinct from traditional NNs, GNNs combine the irregular communication-intensive patterns of graph processing and the regular computation-intensive patterns of NNs. This feature can lead to ineffectual computation on CPUs and GPUs. 
(2)~Since these frameworks assemble functions in a sequential way, one function will not start until the previous one finishes. This execution model leads to extra memory accesses, footprint, and implicit barriers for intermediate results, limiting the potential performance, energy consumption and the scale of graph datasets.

Field-Programmable Gate Arrays (FPGAs) 
are potentially an attractive approach to 
\add{GNN inference acceleration.}
FPGAs' massive fined-grained parallelism provides opportunities to exploit GNNs' inherent parallelism. They also deliver better performance per watt than general-purpose computing platforms. In addition, FPGAs' reconfigurability and concurrency provide great flexibility to solve the challenges of hybrid computing patterns and ineffectual execution. 
\add{Most of the prior works investigating FPGAs focus on 
accelerating a specific GNN model implemented using Hardware Description Languages (HDL).
AWB-GCN~\cite{geng2020awb}, as one of the eariest FPGA-based works, proposes a GCN accelerator using HDL to solve the workload imbalance problem due to the distinct sparsity of different components.
BoostGCN~\cite{zhang2021boostgcn} proposes a graph partition algorithm in a preproessing step to address workload imbalance issues.}
\add{Despite these promising results, HDL design methodology}
is not suitable for widespread adoption for GNN implementations due to the conflict between the non-trivial development efforts with HDL and the rapid emergence of new GNN models. 
To address this challenge, High-Level Synthesis (HLS) tools are proposed to 
\add{create GNN kernels}
using popular languages such as C/C++. 
With the help of HLS, development time is substantially shortened relative to HDL designs.
\add{Lin et al.~\cite{lin2021gcn}, as one of the first works, proposes an HLS-based accelerator for GCN with separated sparse-dense matrix multiplication units and dense matrix multiplication units which are connected by shared memory and execute sequentially. 
GenGNN~\cite{abi2022gengnn} proposes a framework to accelerate GNNs for real-time requirements where the whole graph and corresponding intermediate results are stored in on-chip resources on the FPGA. Despite these promising results, this work is limited to small-scale graphs with low edge-to-node ratio due to on-chip memory usage being proportional to graph scale and feature dimensions.
}

Distinct from pure software programming, HLS developers need to adopt multiple optimization pragmas and follow certain coding styles to achieve best performance and energy cost. 
As reported in~\cite{brown2020exploring}, the performance difference between a well-optimized version and a non-optimized version of the same kernel can be two to three orders of magnitude. 
This invites an open question: \emph{how effectively can modern HLS tools accelerate GNN inference?} 


In this paper, we introduce GNNHLS\footnote{Released 
as a benchmark suite~\cite{gnnhls} and also available at \url{https://github.com/ChenfengZhao/GNNHLS}
}
\add{an open-source framework for comprehensive evaluation of GNN kernels on FPGAs via HLS.}
GNNHLS contains a software stack extended from 
\add{a prior} GNN benchmark~\cite{dwivedi2020benchmarking}
based on PyTorch and DGL for input data generation and conventional platform baseline deployments (i.e., CPUs and GPUs).
\add{It also contains six well-optimized general-purpose GNN applications.} 
\add{These kernels can be classified into 2 classes:}
(1) isotropic GNNs in which every neighbor contributes equally to the update of the target vertex, and (2) anisotropic GNNs in which edges and neighbors contribute differently to the update due to the adoption of operations such as attention and gating mechanisms. In this paper, we make several contributions:
\begin{itemize}
    \item We propose GNNHLS, a framework to evaluate GNN inference acceleration via HLS, containing: (a)~a 
    software stack \add{based on PyTorch and DGL} for data generation and baseline deployment, and (b)~FPGA implementation including 6 well-tuned GNN HLS kernels with host and configuration files \add{which can also be used as benchmarks}. 
    \item We characterize the GNN kernels in terms of locality scores and instruction mix to obtain insight into their memory access and computational properties.
    \item We provide a comprehensive evaluation of our GNN HLS implementations \add{on 4 graph datasets}, assessing both performance improvement and energy reduction. 
\end{itemize}

Our evaluation results show that GNNHLS provides up to $50.8\times$ speedup and $423\times$ energy reduction relative to the multicore CPU baseline. 
\add{Compared with the GPU baselines, GNNHLS achieves up to $5.16\times$ speedup and $74.5\times$ energy reduction.}

\section{Framework Description}
\label{sec:bench_describe}
\subsection{GNNHLS Overview}
\label{sec:bench_overview}

\begin{figure}[tb]
  \centering
  \includegraphics[width=0.98\linewidth]{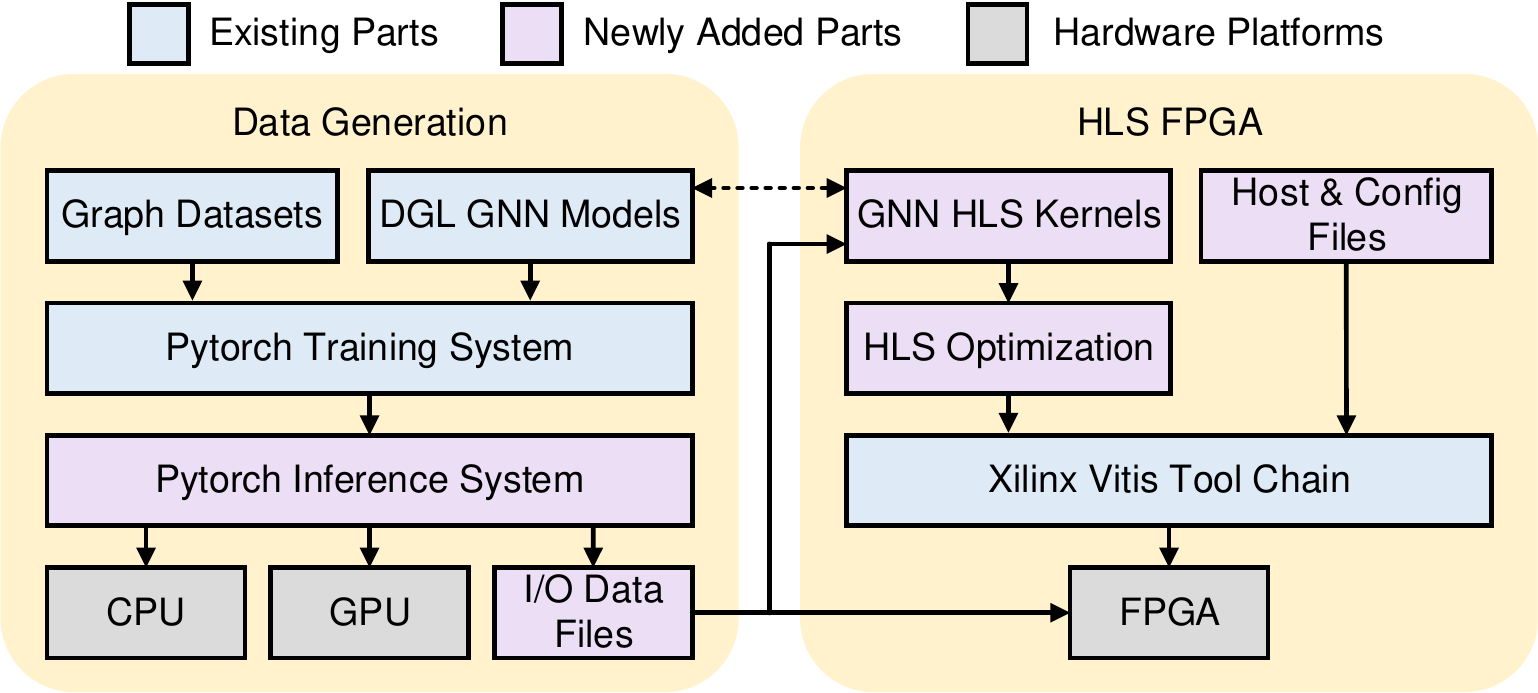}
  \caption{\add{Diagram of} the GNNHLS framework. 
  } 
  \label{fig:bench_framework}
\end{figure}

\add{The GNNHLS framework},
as depicted in Figure~\ref{fig:bench_framework}, comprises two primary components: data generation and HLS FPGA. The former is designed to generate input and output files and measure baselines on a CPU and a GPU, while the latter is designed to implement the optimized HLS applications on an FPGA. The data generation component mainly consists of the training system and the inference system, which are based on PyTorch and DGL. To account for the impact of graph topology on GNN model performance, it uses graph datasets with various topologies, 
including those from Open Graph Benchmarks~\cite{hu2020open}. In addition, six commonly used DGL GNN models obtained from a previous GNN benchmark~\cite{dwivedi2020benchmarking} are incorporated. Thus, realistic model parameters, generated in the training phase, are utilized in inference. 

The HLS FPGA component implements the GNN
kernels on the FPGA. 
These kenels match the functionality of the 
DGL 
baselines \add{and are optimized with several optimization techniques~\cite{de2020transformations}.
}
The optimized HLS kernels, with associated host files, data header files, and configuration files, are compiled by Vitis and executed on the FPGA. 
The optimization techniques applied \add{in GNNHLS} are described as follows: 

\add{\textbf{Pipeline}: Enable instruction-level concurrent execution 
to 
improve overall throughput.
\textbf{Loop Merge}: Optimize the finite state machine (FSM) of nested loops to remove the impact of inner loop latency on the overall throughput. 
\textbf{Burst Memory Access \& Memory Port Widening}: access large chunks of data in contiguous addresses and increase memory port width to improve memory bandwidth. 
\textbf{Loop Unroll}: Leverage instruction-level parallelism by executing multiple copies of loop iterations in parallel to increase throughput 
at the cost of resource utilization. 
\textbf{Dataflow}: 
Enable task-level parallelism by connecting multiple functions with FIFOs to form a pipeline-style architecture and executing them concurrently.
\textbf{Multiple Compute Units (CUs)}: Execute multiple kernel instances as CUs in parallel for different data portions at the cost of resource usage. 
}

\begin{figure}[tb]
  \centering
  \includegraphics[width=0.98\linewidth]{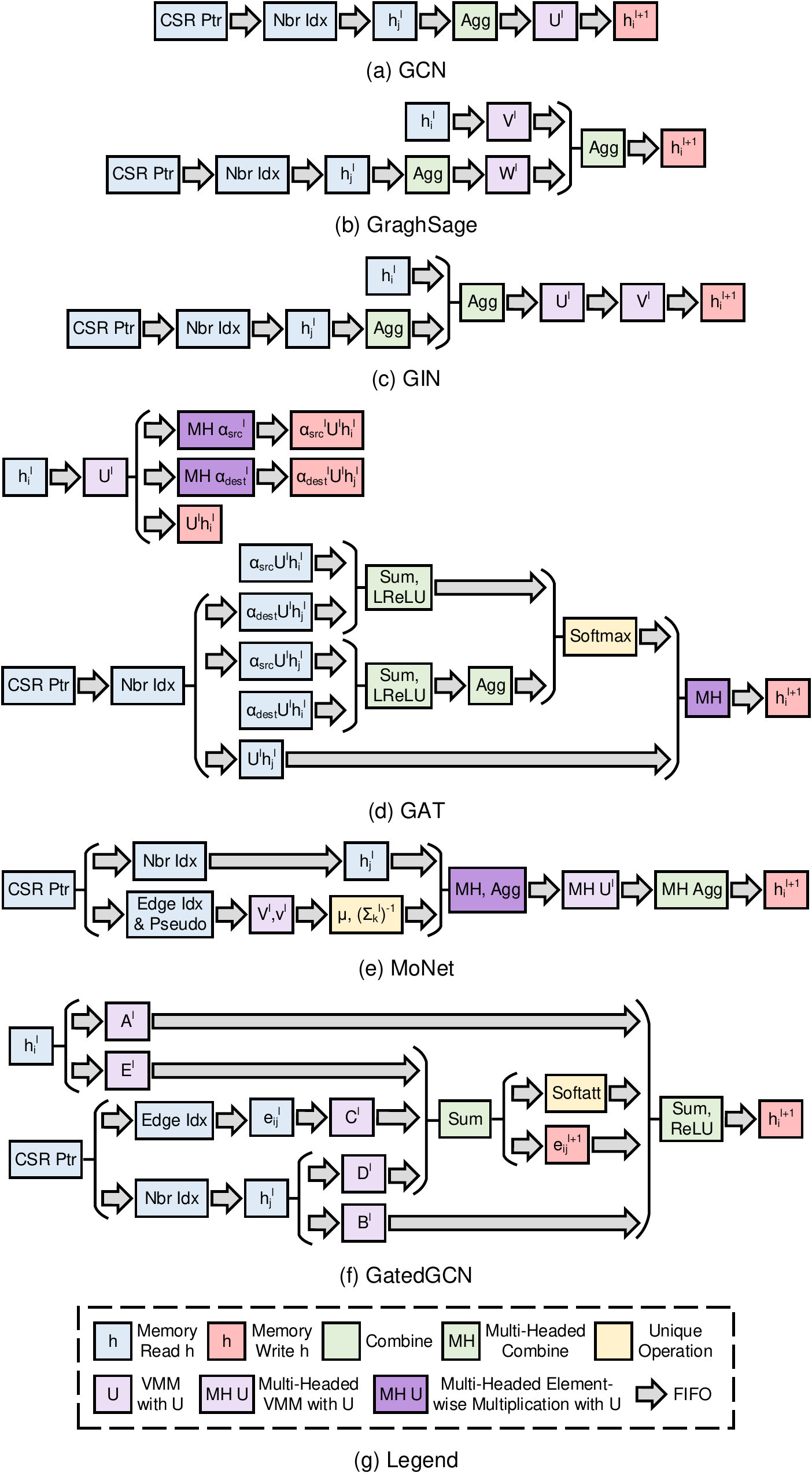}
  \caption{\add{Dataflow diagrams of GNN HLS kernels in GNNHLS.}}
  \label{fig:gnn_diagram}
\end{figure}

Figure~\ref{fig:gnn_diagram} illustrates the Dataflow diagrams of the GNNHLS kernels\add{, in which memory and computation operations are divided and pipelined based on the complexity of each kernel.} 
To mitigate the cost of Dataflow, we also (1) tune the location of FIFO accesses to achieve better throughput, (2) apply vectors for FIFO widening and associated operations, and (3) split loops to optimize the FIFO properties of loop indices.


\subsection{Graph Convolutional Network (GCN)}


Graph Convolutional Network (GCN)~\cite{kipf2016semi} is one of the earliest GNN models and has a simple structure. It updates node features by aggregating 
neighboring node features
and performing linear projection. 
The formula is given as follows:
\begin{equation}
\begin{aligned}
    h_i^{l+1} = \mathrm{ReLU}\left(U^l\sum_{j \in N_i}h_j^l\right)
\end{aligned}
\end{equation}
Where $U^l \in \mathbb{R}^{d \times d}$ is the learnable weight matrix of the linear projection, which performs vector-matrix multiplication. $h_i^l \in \mathbb{R}^{d \times 1}$ is the feature vector of vertex $i$ in layer $l$, and $N_i$ represents the neighboring vertices of vertex $i$.


Based on the above equation, we create the GCN HLS implementation, the Dataflow diagram of which is depicted in Figure~\ref{fig:gnn_diagram}(a).
In addition to the memory access modules for input graphs and $h$, we split the computation operations into two modules: Aggregation of neighbor node vectors $h_j$ and vector-matrix multiplication (VMM) for linear projection. 
We perform all the optimization techniques described previously to the GCN kernel. The memory burst length vector $h$ is $d$, limited by the irregularity of the graph topology. The initiation interval (II) of the aggregation module is $4 \left | N_i \right | + 2$. Since Vitis is not good at synthesizing tree-structured floating-point operations, we separate VMM into 2 functions in the Dataflow scope for grouped VMM and sum, respectively. The II of VMM is thereby reduced from $d^2$ to $d+36$.
All these modules are reused in the following GNN models.
Due to its simplicity, we create 2 CUs to process distinct vertices in parallel. 



\subsection{GraphSage (GS)}


\addvone{GraphSage (GS)~\cite{hamilton2017inductive} introduces an inductive framework to improve the scalability over GCN by aggregating information from the fixed-size set of neighbors via uniform sampling, explicitly incorporating feature vectors of both the target vertex and its source neighbors.}
The mathematical expression of GraphSage \addvone{with a mean aggregator} is formulated as follows:
\begin{equation}
\begin{aligned}
    h_i^{l+1} &= \mathrm{ReLU}\left(U^l \mathrm{Concat}\left(h_i^l, \frac{1}{\left | N_i \right |}\sum_{j \in N_i}h_j^l\right)\right)\\
    &= \mathrm{ReLU}\left(V^l h_i^l + W^l\frac{1}{\left | N_i \right |}\sum_{j \in N_i}h_j^l\right)
\end{aligned}
\end{equation}
Where 
\addvone{$N_i$ is the set of source neighbors of vertex $i$, and $h_i^l \in \mathbb{R}^{d \times 1}$ is the feature vector of vertex $i$ in layer $l$.}
The learnable weight matrix of the linear projection, $U^l \in \mathbb{R}^{d \times 2d}$, is stored in on-chip memory. Given that distinct weight parameters are used for the target vertex and source neighbors, $U^l$ is divided into $V^l \in \mathbb{R}^{d \times d}$ and $W^l \in \mathbb{R}^{d \times d}$, enabling parallel execution of both paths to hide the latency of linear projection for the target vertex.
Figure~\ref{fig:gnn_diagram}(b) illustrates the Dataflow structure of GraphSage. 
The memory read accesses and linear projection of the target feature, and 
neighbors' feature aggregation
are executed simultaneously, and then summed up to update $h_i$. 

\subsection{Graph Isomorphism Network (GIN)}


Graph Isomorphism Network (GIN)~\cite{xu2018how} employs the Weisfeiler-Lehman Isomorphism Test~\cite{weisfeiler1968reduction} as its foundation to investigate the discriminative ability of GNNs. The formula of GIN is described as follows:
\begin{equation}
\begin{aligned}
    h_i^{l+1} = \mathrm{ReLU}\left(U^l \mathrm{ReLU}\left(V^l \left((1+\epsilon)h_i^l + \sum_{j \in N_i}h_j^l\right)\right)\right)
\end{aligned}
\end{equation}
where $\epsilon$ is a learnable scalar weight, $U^l$ and $V^l \in \mathbb{R}^{d \times d}$ denote learnable weight matrices of cascaded VMM modules, $h_i^l \in \mathbb{R}^{d \times 1}$ again refers to the feature vector of vertex $i$ in layer $l$, and $N_i$ is again the source neighbors of vertex $i$. 
In contrast to GraphSage, GIN illustrated in Figure~\ref{fig:gnn_diagram}(c) first sums up the aggregated vector of neighbors $h_j$ and 
the target vertex vector $h_i$,
hiding the latency of reading $h_i$, then performs two cascaded VMM modules with weight matrices $U^l$ and $V^l$, respectively. This framework avoids the generation of long critical paths and achieves a higher clock frequency.

\subsection{Graph Attention Network (GAT)}
\label{sec:gat}


Graph Attention Network (GAT)~\cite{velivckovic2017graph} is an anisotopic GNN model that uses self-attention mechanisms to weight and learn representations of neighbor vertices unequally. The equation is described as follows:

\begin{align}
    h_i^{l+1} &= \mathrm{Concat}_{k=1}^K\left(\mathrm{ELU}\left(\sum_{j \in N_i}  \alpha_{ij}^{k, l} U^{k,l} h_j^l\right)\right) \\
    \alpha_{ij}^{k,l} &= \mathrm{Softmax}(e_{ij}^{k,l}) = \frac{\mathrm{exp}(e_{ij}^{k,l})}{\sum_{j' \in N_i} \mathrm{exp}(e_{ij'}^{k,l})} \\
    e_{ij}^{k,l} &= \mathrm{LeakyReLU}(\Vec{a}^T Concat(U^{k,l} h_i^l, U^{k,l} h_j^l)) \nonumber \\
        &= \mathrm{LeakyReLU}(a_{src}^{k,l} U^{k,l} h_i^l + a_{dest}^{k,l} U^{k,l} h_j^l) 
\end{align}
where $\alpha_{ij}^{l} \in \mathbb{R}^K$ is the attention score between vertex $i$ and vertex $j$ of layer $l$, $U^{k,l} \in \mathbb{R}^{d \times d}$ and $\Vec{a} \in \mathbb{R}^{2d}$ are learnable parameters. Note that  the weight parameter $\Vec{a}^T$ is decomposed into $a_{src}^l$ and $a_{dest}^l \in \mathbb{R}^d$ in the DGL library, because
it is more efficient in terms of performance and memory footprint by transferring VMM between $U^{k,l}$ and $h^l$ from edge-wise to node-wise operations, especially for sparse graphs where the edge number is larger than the vertex number. 



Figure~\ref{fig:gnn_diagram}(d) depicts the Dataflow framework of GAT. Due to the unbalanced workload of the numerator and the denominator in~(5), the results of $\exp(e_{ij})$, size $O(\left | N_i \right | )$, need to be temporarily stored 
prior to being accumulated. Considering the irregularity and large maximum $\left | N_i \right |$ of graphs, we divide the GAT model into 2 HLS kernels linked to the same memory banks for shared intermediate results: kernel 1 is designed to perform VMM with $U$ and $h$, and multi-headed element-wise multiplication (MHEWM) with $a_{src}$ and $a_{dest}$, respectively, in~(6). After being optimized, the II of MHEWM is $k+112$.
The intermediate results are written back to memory and then read by kernel 2 to implement~(4) and~(5).
Note that $e_{ij}$ is computed twice in parallel to avoid performance degradation and deadlock issues.
The II of aggregation, softmax, and MHEWM  is $k\cdot\left | N_i \right |+2k+38$, $k\cdot\left | N_i \right |+k+17$, and $k\cdot\left | N_i \right |+k+14$, respectively.

\subsection{Mixture Model Networks (MoNet)}


Mixture Model Networks (MoNet)~\cite{monti2017geometric} is a general anisotopic GNN framework designed for graph and node classification tasks using Baysian Gaussian Mixture Model (GMM)~\cite{dempster1977maximum}. The model is formulated as follow:

\begin{align}
    h_i^{l+1} &= \mathrm{ReLU}\left(\sum_{k=1}^K \sum_{j \in N_i} w_k(u_{ij}) U^{k,l} h_j^l\right) \nonumber \\
            &= \mathrm{ReLU}\left(\sum_{k=1}^K U^{k,l} \sum_{j \in N_i} w_k(u_{ij}) h_j^l\right) \\
    w_k(u_{ij}) &= \mathrm{exp}\left(-\frac{1}{2}(u_{ij}^l - \mu_k^l)^T ({\textstyle \sum_{k}^{l}})^{-1} (u_{ij}^l - \mu_k^l)\right) \\
    u_{ij}^l &= \mathrm{Tanh}(V^l pseudo_{ij}^l + v^l) \\
    pseudo_{ij}^l &= \mathrm{Concat}(deg_i^{-0.5}, deg_j^{0.5})
\end{align}
where $v^l \in \mathbb{R}^{2}$, $V^l \in \mathbb{R}^{2 \times 2}$, $\mu \in \mathbb{R}^{K \times 2}$, $({\textstyle \sum_{k}^{l}})^{-1} \in \mathbb{R}^{K \times 2}$, and $U^{l} \in \mathbb{R}^{d \times d}$ are learnable parameters of GMM. $v^l$ and $V^l$ represent the pseudo-coordinates between the target vertex and its neighbors, $\mu \in \mathbb{R}^{K \times 2}$ and $({\textstyle \sum_{k}^{l}})^{-1} \in \mathbb{R}^{K \times 2}$ denote the mean vector and covariance matrix. $U^{k,l}$ is the weight matrix.


The Dataflow diagram of MoNet is depicted in Figure~\ref{fig:gnn_diagram}(e). In our HLS implementation, $pseudo_{ij}$ of each edge is processed by a small VMM module with $V^l$ and $v^l$ in~(9) and the Gaussian Weight Computation module with $\mu$ and $({\textstyle \sum_{k}^{l}})^{-1}$ in~(8). Meanwhile, $h_j$ is read from memory for the subsequent MHEWM with aggregation, MHVMM with $U$, and MH Aggregation modules. Note that we perform the MH VMM with $U$ after aggregation in~(7), transferring it from an edge-wise to node-wise operation to reduce its occurrence.
After optimization, the II of the VMM for $u_{ij}$, Gaussian computation, MHEWM with aggregation, MHVMM with $U$, and MH Aggregation are 1, 1, 4, $d+k+28$, and $7k+10$, respectively. We create 2 CUs for the HLS kernel to process vertices with distinct indices.

\subsection{Gated Graph ConvNet (GatedGCN)}


The Gated Graph ConvNet (GatedGCN)~\cite{bresson2017residual} is a type of anisotropic graph neural network (GNN) model that employs a gating mechanism to regulate the flow of information during message passing, allowing the model to emphasize relevant information and filter out irrelevant one. The gating mechanism utilizes gate functions (e.g., sigmoid) to control the flow of messages at each layer. The mathematical expression for GatedGCN is provided below:
\begin{align}
    h_i^{l+1} &= \mathrm{ReLU}\left(A^l h_i^l + \frac{\sum_{j' \in N_i} B^l h^l_{j'} \odot \sigma(e_{ij'}^{l+1})}{\sum_{j' \in N_i} \sigma(e_{ij'}^{l+1}) + \epsilon}\right) \\
    e_{ij}^{l+1} &= E^l h_i^l + D^l h_j^l + C^l e_{ij}^l
\end{align}
where $A^l$, $B^l$, $D^l$, $E^l$ and $C^l \in \mathbb{R}^{d \times d}$ are learnable matrix parameters, $e_{ij}^l \in \mathbb{R}^{1 \times d}$ denote the edge features from vertex $i$ to $j$ layer $l$, $h_i^l$ represents node features of vertex $i$ in layer $l$, $\odot$ denotes Hadamard product, $\sigma$ denotes the sigmoid function, and $\epsilon$ is a constant for numerical stability. 

Since the soft attention of GatedGCN shown in~(11) is distinct from GAT, performing accumulation operations for $e_{ij}$ on both the numerator and denominator, we implement a single pipeline to build the HLS kernel. 
Figure~\ref{fig:gnn_diagram}(f) illustrates the Dataflow framework of GatedGCN.
To hide the latency of multiple VMM modules in GatedGCN, we perform all of them in parallel with parameters $A$, $B$, $D$, $E$, and $C$, respectively. Then the soft attention module is implemented to update $h_i$.
After optimization, the II of the soft attention and sum modules to generate $h_i^{l+1}$ are $10\cdot\left | N_i \right | + 72$ and 31, respectively.

\section{Experimental Methodology}

\noindent
\textbf{Datasets:} 
\addvtwo{Table~\ref{tab:graph} shows the graph datasets used in our evaluation. All these graphs are collected from Open Graph Benchmark~\cite{hu2020open}, a widely-used graph library for GNNs, and have a wide range of fields and scales. These graphs represent two classes of graphs with distinct topologies used in the GNN community: MH and MT consist of multiple small dense graphs, while AX and PT each consist of one single sparse graph. The maximum and average degree shown in Table~\ref{tab:graph} indicates their varying distributions ranging from regular-like to powerlaw-like.}
\addvtwo{In addition, we set feature dimensions for the kernels: GCN, GraphSage, and GIN have the same input and output dimensions at 128. 
The input, head, and output dimensions of GAT and MoNet are (128, 8, 16) and (64, 2, 64), respectively. 
All the dimensions of GatedGCN are 32.
}

\begin{table}[tb]
\caption{Graph datasets.}
\label{tab:graph}
\centering
\begin{tabular}{|l|r|r|r|r|}
\hline
Dataset            & Node \# & Edge \#  & Max.     & Avg. \\ 
                   &         &          & Deg.     & Deg.      \\\hline
OGBG-MOLTOX21 (MT) & 145459  & 302190   & 6         & 2.1       \\
\hline
OGBG-MOLHIV (MH)   & 1049163 & 2259376  & 10        & 2.2       \\  \hline
OGBN-ARXIV (AX)    & 169343  & 1166243  & 13155     & 6.9       \\ \hline
OGBN-PROTEINS (PT) & 132534  & 79122504 & 7750      & 597.0     \\ \hline
\end{tabular}
\end{table}

\noindent
\textbf{Evaluation methods:} 
To perform evaluation, 
we use a Xilinx Alveo U280 FPGA card,
provided by the Open Cloud Testbed~\cite{leeser2021fpgas}, to execute the HLS kernels. 
This FPGA card provides 8~GB of HBM2 with 32 memory banks at 460~GB/s total bandwidth, 32~GB of DDR memory at 38~GB/s, and 3~super logic regions (SLRs) with 1205K look-up tables (LUTs), 2478K registers, 1816 BRAMs, and 9020 DSPs. We \addvtwo{adopt} 32-bit floating point as the data format. We use Vitis 2020.2 for synthesis and hardware linkage \addvtwo{with the power-profile option enabled to perform power profiling 
during runtime},
and Vitis Analyzer to view resource utilization, execution time and power consumption.
We compare our HLS implementation with CPU and GPU baselines with PyTorch and the 
\add{highly-optimized}
DGL library. We perform CPU baseline runs on an Intel Xeon Silver 4114 at 2.2~GHz with 10 cores, 20~threads, and 13.75~MB L3 cache. The GPU baseline is implemented on an Nvidia RTX 2080 Ti with 2994 CUDA cores at 1.5~GHz and 8~GB GDDR6 at 448~GB/s total bandwidth. We measure the energy consumption of the CPU and GPU baselines using the same technique as prior work~\cite{lin2021gcn}.

\section{Characterization}
\label{sec:characterization}
To capture insight into the 
properties of 
\add{GNNHLS},
we first characterize the GNN kernels using instruction mix, spatial locality, and temporal locality. 
We use Workload ISA-Independent Characterization (WIICA)~\cite{shao2013isa}, a workload characterization tool, to capture ISA-independent properties by generating and parsing a dynamic trace of runtime information. Due to the limits of disk and processing time, profiling the the full trace is impractical. Thus we use uniform random node sampling~\cite{leskovec2006sampling} 
to select a sequence of 500 nodes for evaluation. 

\subsection{Instruction Mix}

\begin{figure}[tb]
  \centering
  \includegraphics[width=0.9\linewidth]{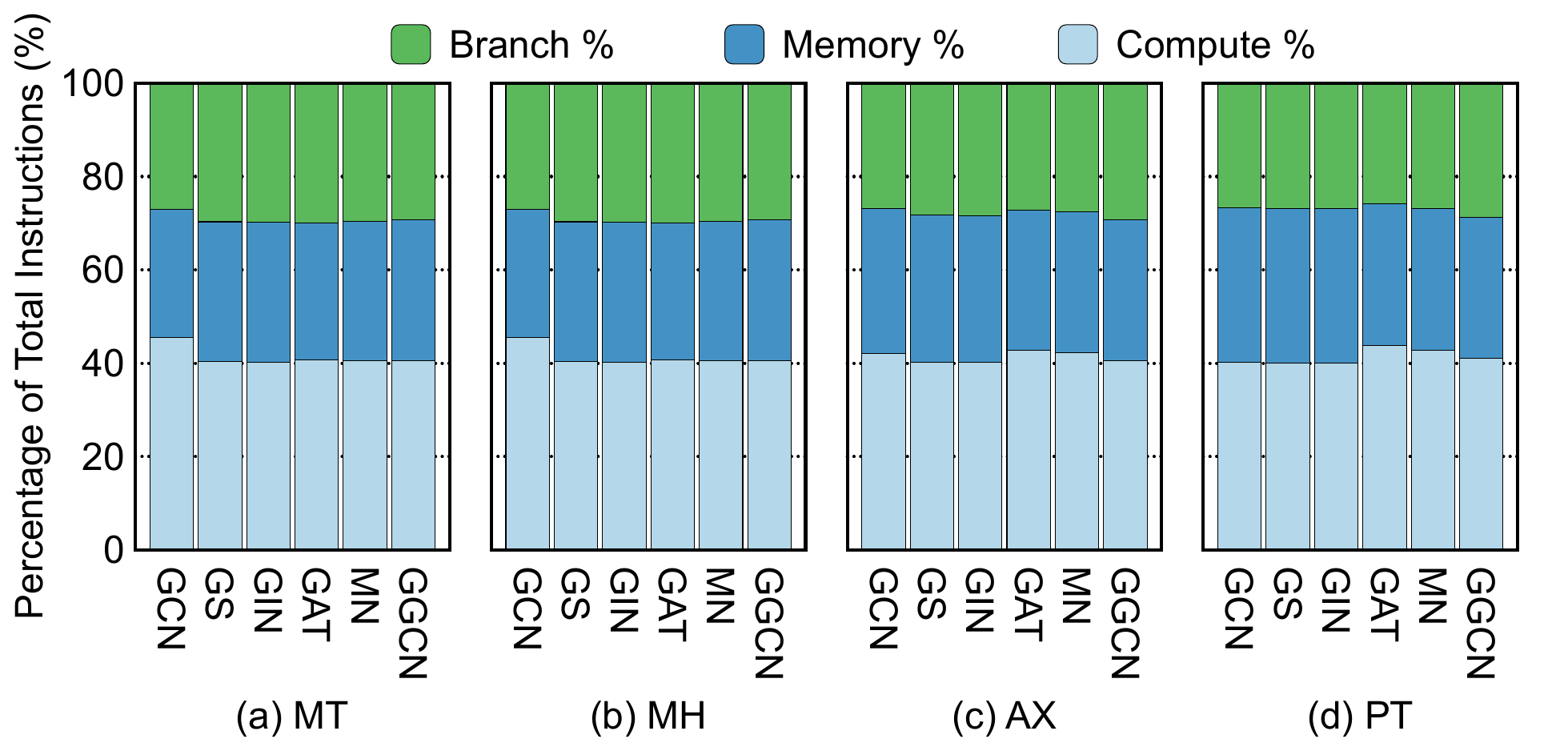}
  \caption{Instruction breakdown of all the HLS kernels.} 
  \label{fig:inst_breakdown}
\end{figure}

We first take a look at the dynamic instruction mix, 
partitioning instructions into 3 classes: branch, memory and compute.
Figure~\ref{fig:inst_breakdown} shows the instruction mix of the HLS kernels on the 4 datasets. 
We observe that the instruction breakdown shows a consistent tendency: (1)~The computation instructions make up largest fraction (about 40\%$-$50\%) of total instructions, implying that these pipeline-style GNN HLS kernels are computation-intensive. (2)~Memory instructions consume the second largest fraction (about 30\%$~$35\%), indicating the total number of memory accesses is still nontrivial even if all the kernels are in a pipeline style. (3)~While branch instructions take 25\%$-$30\% of the total, most of them are due to conditional statements of for loops and irregularity of graphs. We also observe that denser graphs (e.g., AX and PT) induce a higher fraction of compute instructions for anisotropic kernels (i.e., GAT, MN, and GGCN) due to their edge-wise operations. In contrast, denser graphs 
induce a higher fraction of memory instructions for isotropic kernels (i.e., GCN, GS, and GIN) because their edge-wise operations are less computation intensive than node-wise update.

\subsection{Spatial and Temporal Locality}

\begin{figure}[tb]
  \centering
  \includegraphics[width=0.9\linewidth]{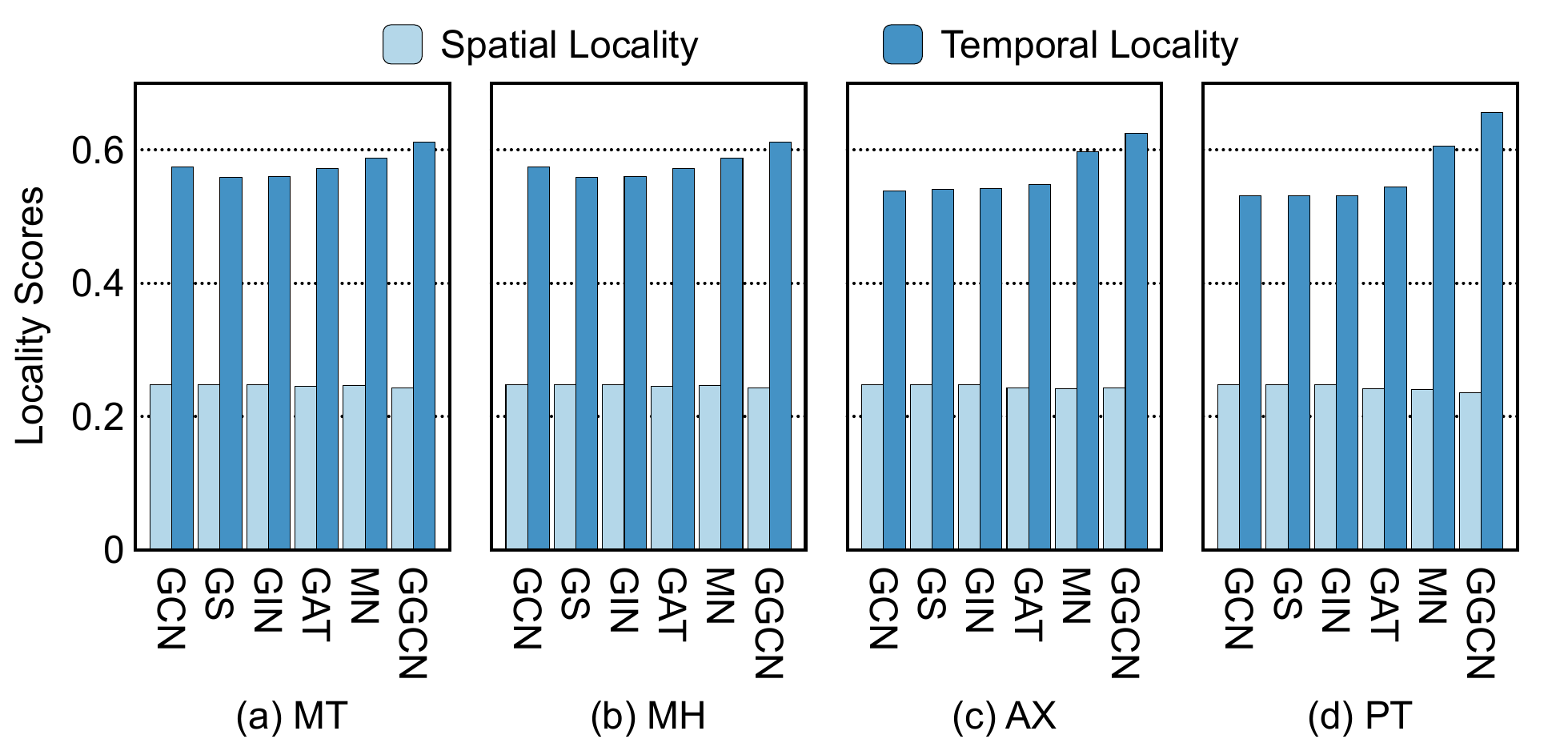}
  \caption{Memory locality scores of HLS kernels.} 
  \label{fig:locality}
\end{figure}

We use spatial locality and temporal locality scores developed by Weinberg et al.~\cite{weinberg2005quantifying} to quantitatively measure the memory access patterns. 
Spatial locality characterizes the closeness of memory references among consecutive memory accesses. For HLS accelerators, it represents the potential opportunity to optimize the efficiency of prefetching and memory burst transfer. 
Temporal locality measures the frequency of memory instructions accessing the same memory address. It represents the latent efficiency of cacheing data elements so that they can be accessed repetitively with lower cost. Therefore, the higher the temporal locality, the more performance improvement due to cacheing mechanisms in the accelerators. 
Both return a score in the range $[0,1]$.

Figure~\ref{fig:locality} illustrates the spatial and temporal locality scores. 
Focusing first on the spatial locality, we observe the score stays consistently low (about $0.23-0.25$) across all the kernels and datasets. It is because the irregularity of graph topology induces non-contiguous memory references, limiting memory burst transfer and prefetching to the length of feature sizes. Next examining the temporal locality, we observe that the score stays in the range of $0.5-0.7$, indicating the potential performance benefit of cacheing mechanisms, 
regardless of the graph topology. In addition, we observe anisotropic kernels show a higher temporal locality than isotropic kernels, due to them having 
more edge-wise operations. 

\section{Evaluation}
\label{sec:eval}

\subsection{Resource Utilization}
\begin{table}[tb]
\caption{Resource Utilization of HLS GNN models.}
\label{tab:resource}
\centering
\begin{tabular}{|l|c|c|c|c|c|c|}
\hline
     & Target      & Actual      & LUT    & FF     & BRAM & DSP  \\
     & Freq.       & Freq.       &        &        &      & \\ \hline
GCN  & 300 MHz    & 250 MHz         & 264485 & 413197 & 41   & 2880 \\ \hline
GS   & 250 MHz    & 204 MHz         & 253608 & 358722 & 33   & 2766 \\ \hline
GIN  & 300 MHz    & 190 MHz         & 278251 & 421915 & 55   & 3264 \\ \hline
GAT  & 300 MHz    & 255 MHz         & 168559 & 248424 & 81   & 1718 \\ \hline
MN   & 300 MHz    & 250 MHz         & 289208 & 428917 & 212  & 2236 \\ \hline
GGCN & 300 MHz    & 270 MHz         & 151497 & 235484 & 124  & 1036 \\ \hline
\end{tabular}
\end{table}

We first 
examine the resource utilization and clock frequency after place \& route. FPGA resources include look-up tables (LUT), flip-flops (FF), BRAM, and digital-signal-processors (DSP). Table~\ref{tab:resource} shows 
these results.
From the table, we observe that the frequency of all the kernels is lower than the target frequency, which is not unusual in FPGA designs. 
Among these kernels, GraphSage achieves a low frequency due to some critical paths which are unresolvable by the tool. In addition, we observe that the resources on the FPGA are not over-utilized.

\begin{figure*}[tb]
  \centering
  \includegraphics[width=0.98\linewidth]{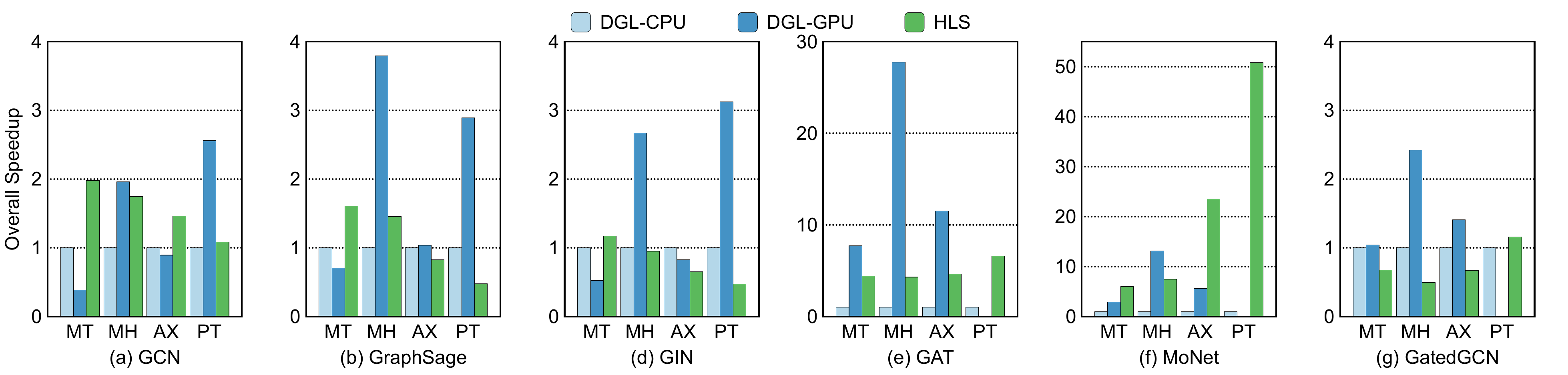}
  \caption{Speedup of HLS kernels relative to DGL-CPU. \add{The higher the better.}} 
  \label{fig:speedup}
\end{figure*}

\begin{table*}[htb]
\caption{Execution time (sec) of DGL-CPU, DLG-GPU, and GNN HLS implementation on 4 graph datasets.}
\label{tab:exec_time}
\centering
\begin{tabular}{|c|ccc|ccc|ccc|ccc|}
\hline
     & \multicolumn{3}{c|}{MT}                                            & \multicolumn{3}{c|}{MH}                                            & \multicolumn{3}{c|}{AX}                                            & \multicolumn{3}{c|}{PT}                                             \\ \hline
     & \multicolumn{1}{c|}{DGL-CPU} & \multicolumn{1}{c|}{DGL-GPU} & HLS  & \multicolumn{1}{c|}{DGL-CPU} & \multicolumn{1}{c|}{DGL-GPU} & HLS  & \multicolumn{1}{c|}{DGL-CPU} & \multicolumn{1}{c|}{DGL-GPU} & HLS  & \multicolumn{1}{c|}{DGL-CPU} & \multicolumn{1}{c|}{DGL-GPU} & HLS   \\ \hline
GCN  & \multicolumn{1}{c|}{0.11}    & \multicolumn{1}{c|}{0.28}    & 0.05 & \multicolumn{1}{c|}{0.69}    & \multicolumn{1}{c|}{0.35}    & 0.39 & \multicolumn{1}{c|}{0.31}    & \multicolumn{1}{c|}{0.34}    & 0.21 & \multicolumn{1}{c|}{16.09}   & \multicolumn{1}{c|}{6.29}    & 14.85 \\ \hline
GS   & \multicolumn{1}{c|}{0.21}    & \multicolumn{1}{c|}{0.30}    & 0.13 & \multicolumn{1}{c|}{1.42}    & \multicolumn{1}{c|}{0.38}    & 0.98 & \multicolumn{1}{c|}{0.43}    & \multicolumn{1}{c|}{0.42}    & 0.52 & \multicolumn{1}{c|}{16.45}   & \multicolumn{1}{c|}{5.68}    & 34.29 \\ \hline
GIN  & \multicolumn{1}{c|}{0.15}    & \multicolumn{1}{c|}{0.29}    & 0.13 & \multicolumn{1}{c|}{0.93}    & \multicolumn{1}{c|}{0.35}    & 0.98 & \multicolumn{1}{c|}{0.34}    & \multicolumn{1}{c|}{0.41}    & 0.52 & \multicolumn{1}{c|}{16.11}   & \multicolumn{1}{c|}{5.15}    & 34.29 \\ \hline
GAT  & \multicolumn{1}{c|}{0.91}    & \multicolumn{1}{c|}{0.12}    & 0.21 & \multicolumn{1}{c|}{6.52}    & \multicolumn{1}{c|}{0.24}    & 1.51 & \multicolumn{1}{c|}{3.10}    & \multicolumn{1}{c|}{0.27}    & 0.67 & \multicolumn{1}{c|}{186.93}  & \multicolumn{1}{c|}{OoM}     & 28.28 \\ \hline
MN   & \multicolumn{1}{c|}{0.32}    & \multicolumn{1}{c|}{0.11}    & 0.05 & \multicolumn{1}{c|}{2.37}    & \multicolumn{1}{c|}{0.18}    & 0.32 & \multicolumn{1}{c|}{1.18}    & \multicolumn{1}{c|}{0.21}    & 0.05 & \multicolumn{1}{c|}{89.71}   & \multicolumn{1}{c|}{OoM}     & 1.77  \\ \hline
GGCN & \multicolumn{1}{c|}{0.12}    & \multicolumn{1}{c|}{0.11}    & 0.17 & \multicolumn{1}{c|}{0.62}    & \multicolumn{1}{c|}{0.26}    & 1.26 & \multicolumn{1}{c|}{0.36}    & \multicolumn{1}{c|}{0.26}    & 0.54 & \multicolumn{1}{c|}{38.93}   & \multicolumn{1}{c|}{OoM}     & 33.55 \\ \hline
\end{tabular}
\end{table*}

\subsection{Performance}
We next examine the performance improvement by showing the overall speedup, defined as the execution time of the 
\add{GNN HLS kernels} 
relative to CPU-DGL (using all 10 cores on the CPU)\add{, in Figure~\ref{fig:speedup}}.
\add{Table~\ref{tab:exec_time} shows the execution time of baselines and HLS kernels. Note that GPU results of GAT, MN, and GGCN on PT cannot be obtained because of running out of memory (OoM).}
Examining 
\add{each kernel in Figure~\ref{fig:speedup}}, 
we observe that the HLS implementation is not always outperforming 
\add{corresponding CPU baselines.}
\add{Compared with DGL-CPU, the speedup ranges from $0.47\times$ to $50.8\times$.}

\add{Among isotropic GNN 
kernels, GCN achieves better performance than GraphSage and GIN, ranging from $1.08\times$ to $1.98\times$ because its simpler structure enables us to create two CUs to leverage spatial data parallelism. In contrast, we can only create one CU for GraphSage and GIN each because of their complex structure and heavy resource usage. In addition, we observe that the execution time of GraphSage and GIN are close. Thus, we conclude that the distinction on the structure of these two GNN models will not substantially affect HLS implementation results. 

Among anisotropic kernels, MoNet achieves highest performance improvement ranging from $6.04\times$ to $50.8\times$ due to (1)~its single pipeline structure with computation order optimization where the node-wise operations are placed behind the edge-wise operations, and (2) well-designed MHVMM modules with lower II, especially MHVMM whose II is $O(d+k)$ instead of $O(dk)$. In spite of the 2-pipeline structure of GAT,
we observe that it still achieves $4.31\times$ to $6.61\times$ speedup relative to multi-core CPU baselines.
In addition, since the feature size of GatedGCN 
is smaller,
leading to more performance improvement for CPU baselines with time complexity of $O(d^2)$,
its speedup is not comparable to other anisotropic kernels, ranging from $0.5\times$ to $1.16\times$.
}

\add{Turning our attention to how the performance benefit of HLS implementations varies across graph datasets, we observe that the speedup of isotropic kernels relative to DGL-CPU on regular-like graphs (i.e., MT and MH) is higher than powerlaw-like graphs (i.e., AX and PT) because (1)~the edge-wise operations are less computation-intensive than node-wise operations in these kernels, 
making the baselines more computationally efficient on powerlaw-like graphs containing more edges than nodes; and (2)~the edge-wise aggregation operations in HLS implementations are executed sequentially without leveraging edge-level parallelism, 
making these HLS kernels less computationally efficient for powerlaw-like graphs.
Distinct from isotropic kernels, the speedup of anisotropic kernels on powerlaw-like graphs is higher than regular-like graphs because the edge-wise operations of these kernels are more computation-intensive than isotropic kernels, making baselines less efficient on powerlaw-like graphs.
}

Focusing on the second and the third bar, we observe that DGL-GPU outperforms HLS implementations in many cases, due to the high-performance fixed-function 
accelerators in the GPU.
The speedup of HLS kernels relative to the GPU baselines ranges from $0.13\times - 5.16\times$. In spite of the promising GPU performance, 
\add{there are still some drawbacks of GPU compared with HLS implementations. For the execution of isotropic GNN models, DGL-GPU achieves lower speedup than HLS on small-scale graphs such as MT and AX.
It is speculated that the GPU is designed to achieve high throughput in the cost of latency which plays a more important role for small-scale graphs than large-scale graphs. 
In addition, compared with HLS implementations on FPGA, GPU is also not suitable for the execution of anisotropic GNN models on large-scale, especially powerlaw-like graphs (e.g., PT) due to (1)~the non-trivial memory footprint caused by its sequential execution paradigm to store intermediate results of edge-wise operations, and (2)~insufficient memory capacity on the GPU board. That is why we failed to execute anisotropic GNNs on PT with GPU. It is solved by the HLS implementations' pipeline structure not storing the intermediate results.
}

\add{Since GenGNN~\cite{abi2022gengnn} also discusses 3 of the GNN models included in this paper (GCN, GIN, and GAT), we can make a limited comparison of our GNN HLS implementations with theirs. The two are not directly comparable for a number of reasons: (1)~the feature dimensions of our GNN HLS kernels are higher, (2)~we use off-chip memory instead of on-chip memory, (3)~our general-purpose GNN HLS kernels focus more on throughput rather than real-time latency, and (4)~the FPGAs are from the same family, but are not same part.
The performance of our HLS kernels exceeds that of GenGNN, achieving overall speedup of $35\times$, $5\times$, and $6\times$ over GCN, GIN, and GAT, on MT, respectively.}

\subsection{Optimization Techniques}
As described in Section~\ref{sec:bench_describe}, we apply multiple optimization techniques 
\add{to the HLS kernels}. 
In order to evaluate the efficacy of these techniques, we use GraphSage on MT as a case study. Table~\ref{tab:opt} presents the execution time of GraphSage with the combined impact of optimization techniques applied. 
The reported execution time of each technique represents the effect of both the current technique and 
\add{above techniques listed in the table.}
In the table, 
No Pragma means we don't intentionally apply any pragmas to the HLS code\add{, except for those automatically applied by Vitis (i.e., Pipeline, Loop Merge, and Memory optimizations).}
Dataflow denotes that we apply dataflow pragma and FIFO streams to exploit the task-level parallelism of each application. Loop Unroll means we apply loop unroll pragmas to completely or partially unroll for loops, keeping II as low as possible while exploiting instruction parallelism. Vectorization means using vector data types to widen the width of FIFO streams 
\add{and corresponding operations to decrease the cost of FIFO accesses.}
Split Loops means splitting the outer-most node loop and putting it inside each function connected by streams 
\add{to further optimize FIFO properties inferred from loop indices.}

We observe that Loop Unroll achieves the highest performance improvement. 
\add{Therefore,} exploiting instruction parallelism is still the primary choice for GNN HLS optimization. In order to further improve performance, exploiting task-level parallelism is necessary. Focusing on the first and second row in the table, we observe that 
\add{only}
performing the dataflow pragma and streams in a naive way obtains $1.99\times$ performance improvement. 
\add{By applying Vectorization and Split Loops as complementary techniques of Dataflow,} performance is further improved by $2.5\times$ and $3.9\times$, respectively. 
After applying all the optimization techniques together we observe that the performance of GraphSage is improved by $132\times$. 

\begin{table}[tb]
\caption{Execution time of various optimization techniques for GraphSage on MH. 
}
\label{tab:opt}
\centering
\begin{tabular}{|l|c|r|}
\hline
Optimizations & Execution Time (s) & Speedup \\ \hline
No Pragmas    & 129.59             & 1.00$\times$    \\ \hline
Dataflow      & 65.11              & 1.99$\times$    \\ \hline
Loop Unroll   & 11.11              & 11.7$\times$   \\ \hline
Vectorization & 4.44               & 29.2$\times$   \\ \hline
Split Loops & 0.98               & 132$\times$  \\ \hline
\end{tabular}
\end{table}

\subsection{Energy Consumption}
\begin{figure*}[tb]
  \centering
  \includegraphics[width=0.98\linewidth]{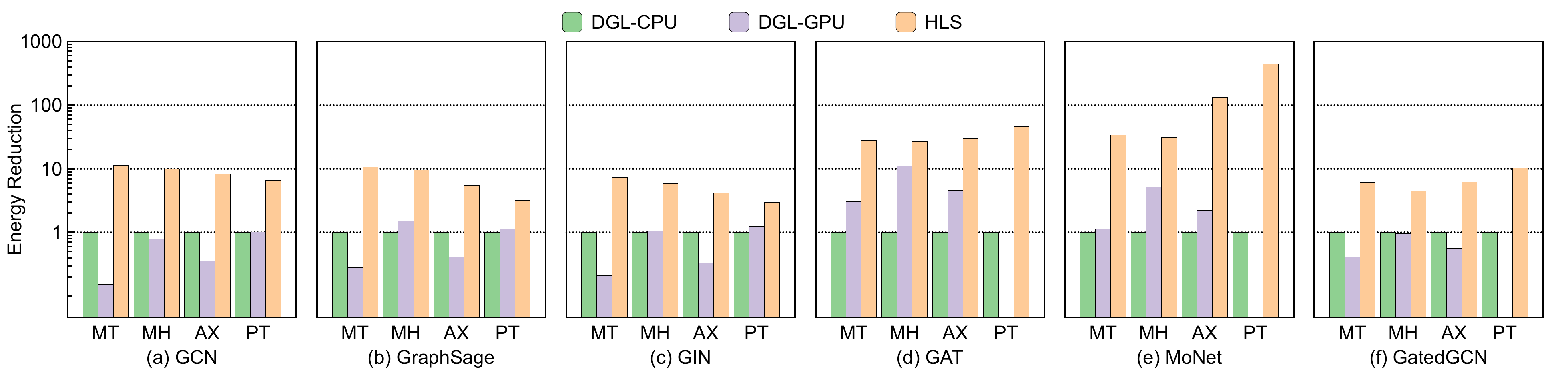}
  \caption{Energy consumption reduction of HLS kernels relative to DGL-CPU (logarithmic scale). \add{The higher the better.}} 
  \label{fig:eng}
\end{figure*}

\begin{table*}[tb]
\caption{Energy Consumption (J) of DGL-CPU, DGL-GPU, and GNN HLS implementation on 4 graph datasets.}
\label{tab:eng}
\centering
\begin{tabular}{|c|ccc|ccc|ccc|ccc|}
\hline
     & \multicolumn{3}{c|}{MT}                                            & \multicolumn{3}{c|}{MH}                                             & \multicolumn{3}{c|}{AX}                                            & \multicolumn{3}{c|}{PT}                                               \\ \hline
     & \multicolumn{1}{c|}{DGL-CPU} & \multicolumn{1}{c|}{DGL-GPU} & HLS  & \multicolumn{1}{c|}{DGL-CPU} & \multicolumn{1}{c|}{DGL-GPU} & HLS   & \multicolumn{1}{c|}{DGL-CPU} & \multicolumn{1}{c|}{DGL-GPU} & HLS  & \multicolumn{1}{c|}{DGL-CPU}  & \multicolumn{1}{c|}{DGL-GPU} & HLS    \\ \hline
GCN  & \multicolumn{1}{c|}{9.06}    & \multicolumn{1}{c|}{59.67}   & 0.80 & \multicolumn{1}{c|}{58.38}   & \multicolumn{1}{c|}{75.25}   & 5.85  & \multicolumn{1}{c|}{25.93}   & \multicolumn{1}{c|}{73.38}   & 3.10 & \multicolumn{1}{c|}{1367.75}  & \multicolumn{1}{c|}{1352.67} & 208.77 \\ \hline
GS   & \multicolumn{1}{c|}{17.95}   & \multicolumn{1}{c|}{64.60}   & 1.68 & \multicolumn{1}{c|}{120.97}  & \multicolumn{1}{c|}{80.63}   & 12.73 & \multicolumn{1}{c|}{36.74}   & \multicolumn{1}{c|}{89.54}   & 6.69 & \multicolumn{1}{c|}{1397.99}  & \multicolumn{1}{c|}{1221.69} & 439.91 \\ \hline
GIN  & \multicolumn{1}{c|}{13.12}   & \multicolumn{1}{c|}{63.20}   & 1.77 & \multicolumn{1}{c|}{79.25}   & \multicolumn{1}{c|}{75.04}   & 13.40 & \multicolumn{1}{c|}{29.10}   & \multicolumn{1}{c|}{89.11}   & 7.10 & \multicolumn{1}{c|}{1369.04}  & \multicolumn{1}{c|}{1107.06} & 464.29 \\ \hline
GAT  & \multicolumn{1}{c|}{77.45}   & \multicolumn{1}{c|}{25.37}   & 2.79 & \multicolumn{1}{c|}{554.10}  & \multicolumn{1}{c|}{50.53}   & 20.50 & \multicolumn{1}{c|}{263.09}  & \multicolumn{1}{c|}{57.74}   & 8.83 & \multicolumn{1}{c|}{15889.04} & \multicolumn{1}{c|}{OoM}     & 344.14 \\ \hline
MN   & \multicolumn{1}{c|}{27.46}   & \multicolumn{1}{c|}{24.32}   & 0.80 & \multicolumn{1}{c|}{201.19}  & \multicolumn{1}{c|}{38.70}   & 6.48  & \multicolumn{1}{c|}{100.59}  & \multicolumn{1}{c|}{45.59}   & 0.75 & \multicolumn{1}{c|}{7625.48}  & \multicolumn{1}{c|}{OoM}     & 17.22  \\ \hline
GGCN & \multicolumn{1}{c|}{9.84}    & \multicolumn{1}{c|}{23.82}   & 1.62 & \multicolumn{1}{c|}{53.12}   & \multicolumn{1}{c|}{55.47}   & 12.05 & \multicolumn{1}{c|}{30.76}   & \multicolumn{1}{c|}{55.32}   & 5.00 & \multicolumn{1}{c|}{3309.16}  & \multicolumn{1}{c|}{OoM}     & 323.44 \\ \hline
\end{tabular}
\end{table*}

We next present a quantitative analysis of the energy consumption. 
Figure~\ref{fig:eng} displays the energy reduction of both DGL-GPU and HLS implementations relative to DGL-CPU in logarithmic scale. Energy reduction is calculated as the energy consumption of DGL-GPU or HLS divided by that of DGL-CPU. Examining the final bar of each application and dataset, we observe that HLS implementations consume less energy than 
\add{CPU and GPU baselines} in all cases. The energy reduction ranges from $2.95\times$ to $423\times$ relative to DGL-CPU and from $2.38\times$ to $74.5\times$ relative to DGL-GPU.
It is because of the low power of FPGA logic, low clock frequency, and efficient pipeline structure of HLS implementations. 

Focusing on the first and last bar, 
\add{we observe a similar tendency in energy reduction as in performance:}
for isotropic GNN models, denser graphs result in lower energy reduction, whereas for anisotropic GNN models, denser graphs result in higher energy reduction. 
This leads us to conclude that improving GNN applications generally will require some degree of graph topology awareness. 

\section{Conclusions}
In this paper, we propose GNNHLS, an open-source framework to comprehensively evaluate GNN inference acceleration on FPGAs via HLS. GNNHLS consists of a software stack for data generation and baseline deployment, and 6 well-tuned GNN HLS kernels. 
We characterize the HLS kernels in terms of instruction mix and memory locality scores, and evaluate them on 4 graph datasets with various topologies and scales. Results show 
up to $50.8\times$ speedup and $423\times$ energy reduction relative to the multi-core CPU baselines. Compared with GPU baselines, GNNHLS achieves up to $5.16\times$ speedup and $74.5\times$ energy reduction.
In the future, we will 
extend GNNHLS to more GNN models and graph datasets. It can also be useful as a benchmark or baseline for HLS researchers to 
explore the potential of HLS tools on GNN inference acceleration. 
GNNHLS has been released for use as a benchmark suite~\cite{gnnhls}.

\section*{Acknowledgment}
This work is supported by NSF under grants CNS-1739643 and CNS-1763503 and a gift from BECS Technology, Inc. The authors are grateful for the use of the Open Cloud Testbed~\cite{leeser2021fpgas} as an experimentation platform.

\bibliographystyle{IEEEtranS}
\IEEEtriggeratref{22}
\bibliography{refs}

\end{document}